\documentclass{article}
\pdfoutput=1
\usepackage{spconf,amsmath,graphicx}
\usepackage{times}
\usepackage{epsfig}
\usepackage{amssymb}
\usepackage{url}

\usepackage{multirow}
\usepackage[multiple]{footmisc}
\usepackage{tabularx,ragged2e,booktabs}
\usepackage{pbox}

\title{VID2SPEECH: SPEECH RECONSTRUCTION FROM SILENT VIDEO}
\name{Ariel Ephrat and Shmuel Peleg
\thanks{This research was supported by Israel Science Foundation, by DFG and by Intel ICRI-CI.}}
\address{The Hebrew University of Jerusalem\\
Jerusalem, Israel}
\begin{document}
%
\maketitle
\begin{abstract}
Speechreading is a notoriously difficult task for humans to perform. In this paper we present an end-to-end model based on a convolutional neural network (CNN) for generating an intelligible acoustic speech signal from silent video frames of a speaking person.
The proposed CNN generates sound features for each frame based on its neighboring frames. Waveforms are then synthesized from the learned speech features to produce intelligible speech.
We show that by leveraging the automatic feature learning capabilities of a CNN, we can obtain state-of-the-art word intelligibility on the GRID dataset, and show promising results for learning out-of-vocabulary (OOV) words. 

\end{abstract}

\section{Introduction}
\label{sec:intro}
\emph{Speechreading} is the task of obtaining reliable phonetic information from a speaker's face during speech perception. It has been described as ``trying to grasp with one sense information meant for another". Given the fact that often several phonemes (phonetic units of speech) correspond to a single viseme (visual unit of speech), it is a notoriously difficult task for humans to perform.

\begin{figure}[t]
    \centering
	\includegraphics[width=\linewidth]{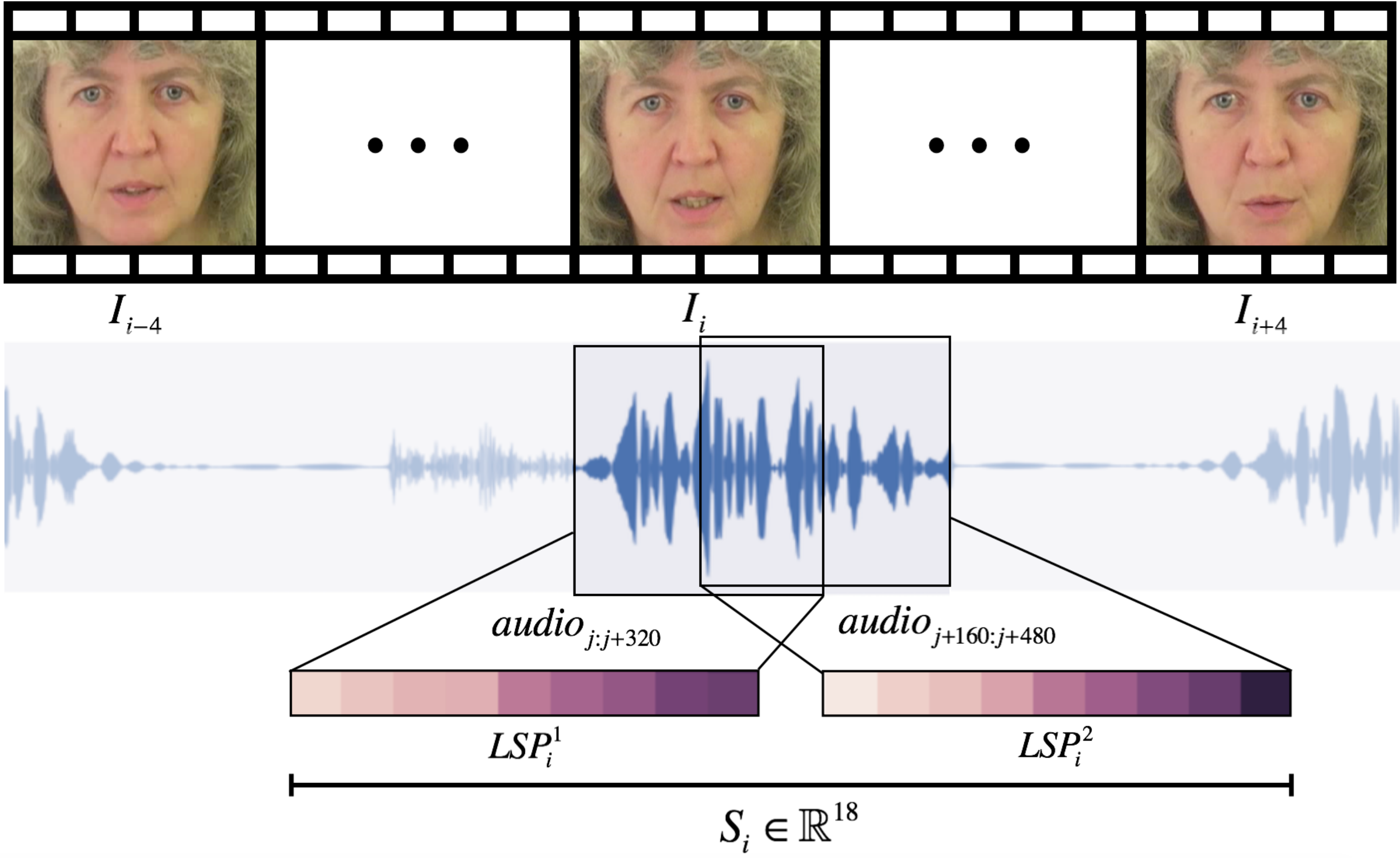}
    \caption{Our CNN-based model takes the frames of silent video as input, and predicts sound features which are converted into intelligible speech. Sound features are calculated by performing $8th$-order LPC analysis and LSP decomposition on half-overlapping audio frames of $40ms$ each. Concatenating every two successive LSP vectors results in a feature vector $S_i \in \mathbb{R}^{18}$.}
    \label{fig:features}
\end{figure}

Several applications come to mind for automatic video-to-speech systems: Enabling videoconferencing from within a noisy environment; facilitating conversation at a party with loud music between people having wearable cameras and earpieces; maybe even using surveillance video as a long-range listening device.

Much work has been done in the area of automating speechreading by computers \cite{petajan1984automatic, matthews2002extraction, zhou2014review}. There are two main approaches to this task. The first, and the one most widely attempted in the past, consists of modeling speechreading as a classification problem. In this approach, the input video is manually segmented into short clips which contain either whole words from a predefined dictionary, or parts of words comprising phonemes or visemes \cite{bear2016decoding}. Then, visual features are extracted from the frames and fed to a classifier. Wand \emph{et al.} \cite{wand2016lipreading}, Assael \emph{et al.} \cite{assaellipnet} and Chung \emph{et al.} \cite{chung2016lip} have all recently showed state-of-the-art word and sentence-level classification results  using neural network-based models.

The second approach, and the one used in this work, is to model speechreading as an articulatory-to-acoustic mapping problem in which the ``label" of each short video segment is a corresponding feature vector representing the audio signal. Kello and Plaut \cite{kello2004neural} and Hueber and Bailly \cite{hueber2016} attempted this approach using various sensors to record mouth movements. Le Cornu and Milner \cite{cornu2015reconstructing} took this direction in a recent work where they used hand-crafted visual features to produce intelligible audio.

A major advantage of this model of learning is its non-dependency on a particular segmentation of the input data into words or sub-words. It does not either need to have explicit manually-annotated labels, but rather uses ``natural supervision" \cite{owens2015visually}, in which the prediction target is derived from a natural signal in the world. 
A regression-based model is also vocabulary-agnostic. Given a training set with a large enough representation of the phonemes/visemes of a particular language, it can reconstruct words that are not present in the training set. Classification at the sub-word level can also have the same effect.
Another advantage to this model is its ability to reconstruct the non-textual parts of human speech, e.g. emotion, prosody, etc.

Researchers have spent much time and effort finding visual features which accurately map facial movements to auditory signal. We bypass the need for feature crafting by utilizing CNNs, which have brought significant advances to computer vision in recent years. Given raw visual data as input, our network automatically learns optimal visual features for reconstructing an acoustic signal closest to the original.

In this paper, we: (1) Present an end-to-end CNN-based model that predicts the speech audio signal of a silent video of a person speaking, significantly improving state-of-the-art reconstructed speech intelligibility; (2) demonstrate that allowing the model to learn from the speaker's entire face instead of only the mouth region greatly improves performance; (3) show that modeling speechreading as a regression problem allows us to reconstruct out-of-vocabulary words.

\begin{figure}[t]
    \centering
	\includegraphics[width=\linewidth]{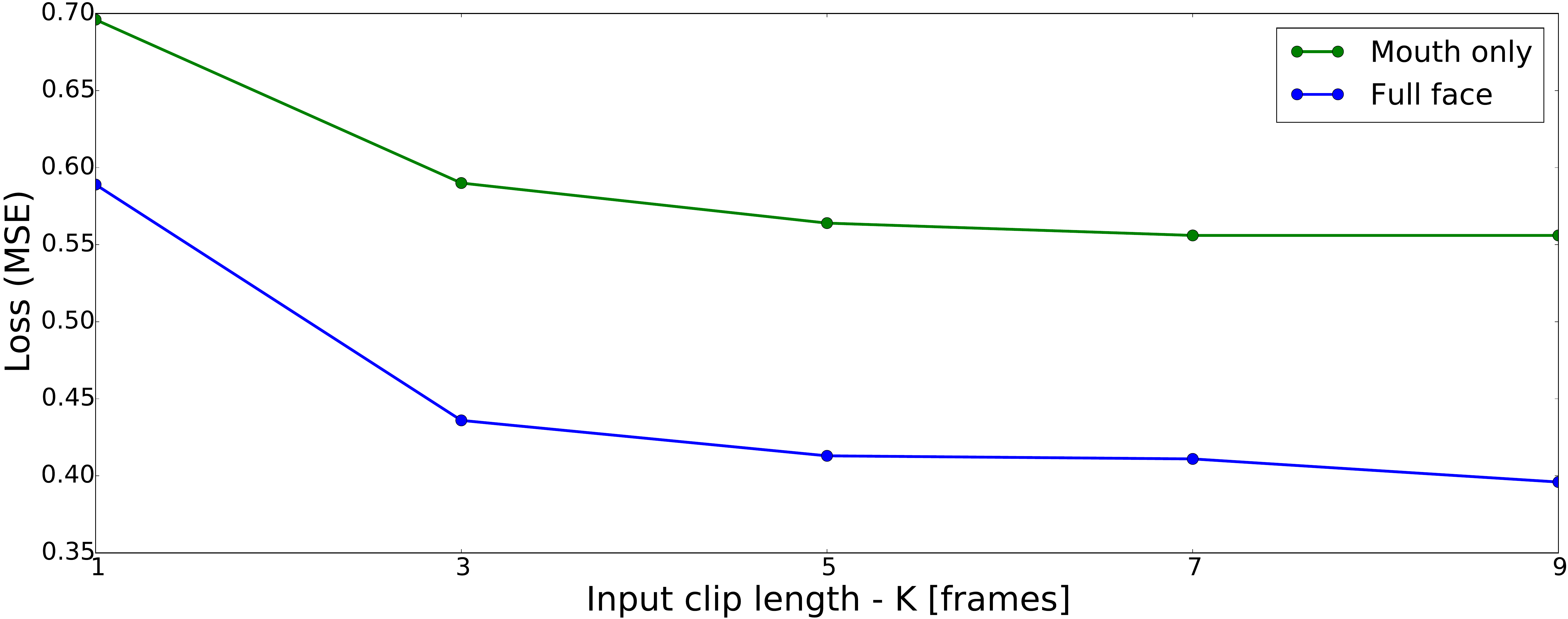}
    \caption{This figure illustrates ($1$) the importance of allowing the network to learn visual features from the speaker's entire face, as opposed to the mouth region only; ($2$) the disambiguation effect of using temporal context. Green line (top) is test error as a function of input clip length $K$ when using only mouth region, and blue line (bottom) is test error when using full face region. Face region error is $40\%$ lower than mouth region in the best configuration.}
    \label{fig:mouth_vs_fullface}
\end{figure}

\section{Speech representation}
\label{sec:speechrep}

The challenge of finding a suitable representation for an acoustic speech signal which can be estimated by a neural network on one hand, and synthesized back into intelligible audio on the other, is not trivial. Spectrogram magnitude, for example, can be used as network output, however the quality of its resynthesis into speech is usually poor, as it does not contain phase information. Use of raw waveform as network output was ruled out for lack of a suitable loss function with which to train the network.

Linear Predictive Coding (LPC) is a powerful and widely used technique for representing the spectral envelope of a digital speech signal, which assumes a source-filter model of speech production \cite{fant1971acoustic}. LPC analysis is applied to overlapping audio frames of the original speech signal, resulting in an LPC coefficient vector whose order \emph{P} can be tuned. Line Spectrum Pairs (LSP) \cite{lsp} are a representation of LPC coefficients which are more stable and robust to quantization and small coefficient deviations. LSPs are therefore useful for speech coding and transmission over a channel, and indeed proved to be well suited to the task at hand.

We apply the following procedure to calculate audio features suitable for use as neural network output: First, the audio from each video sequence is downsampled to $8$kHz and split into audio frames of $40ms$ ($320$ samples) each, with an overlap of $20ms$. $8th$-order LPC analysis is applied to each audio frame, as done by \cite{cornu2015reconstructing}, followed by LSP decomposition, resulting in a feature vector of length $9$ per frame. While $8th$-order LPC is relatively low for high-fidelity modeling of the speech spectrum, we did so in order to isolate the effect of using CNN-learned visual features versus the hand-crafted ones of \cite{cornu2015reconstructing}. Each video frame has two successive corresponding feature vectors, which are concatenated to form a sound vector, $S_i \in \mathbb{R}^{18}$. See Figure \ref{fig:features} for an illustration of this procedure. Finally, the vectors are standardized element-wise by subtracting the mean and dividing by the standard deviation of each element.

\begin{figure*}[t]
    \centering
	\includegraphics[width=\linewidth]{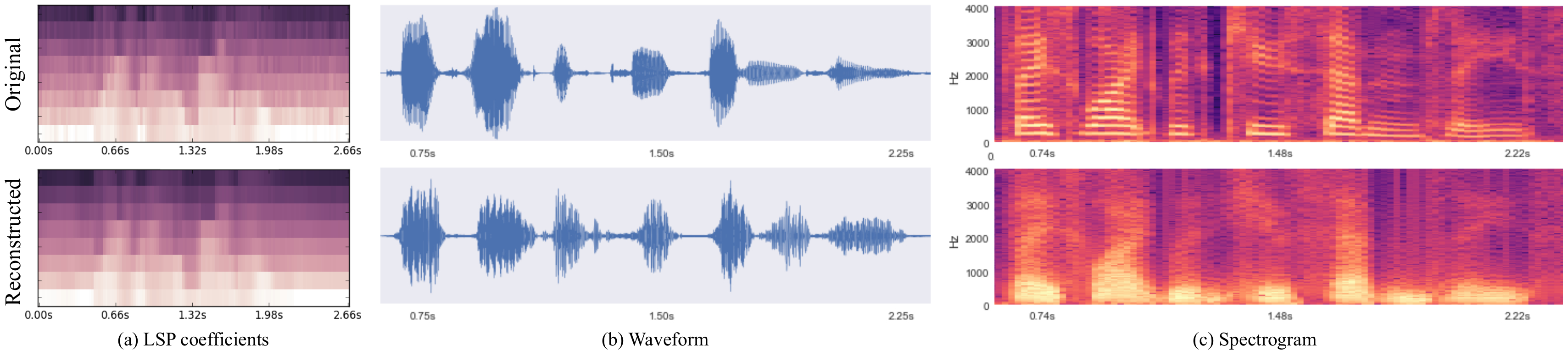}
    \caption{Examples of original (top) and reconstructed (bottom): (a) LSP coefficients, (b) waveform and (c) spectrogram. The vertical columns of (a) are the actual output of the CNN. Spectral envelope of reconstructed audio (c) is relatively accurate, however unvoiced excitation results in the lack of formants (horizontal lines inside spectral envelope, representing frequency of voiced speech).}
    \label{fig:comparison}
\end{figure*}

\section{Predicting speech}
\label{method}

\subsection{Regressing sound features}
\label{ssec:soundfeatures}
Given a sequence of input frames $I_1, I_2, ..., I_N$ we would like to estimate a corresponding sequence of sound features $S_1, S_2, ..., S_N$ where $S_i \in \mathbb{R}^{18}$.

\paragraph*{Input representation}
Our goal is to reconstruct a single audio representation vector $S_i$ which corresponds to the duration of a single video frame $I_i$. However, instantaneous lip movements such as those in isolated video frames can be significantly disambiguated by using a temporal neighborhood as context. Therefore, the input to our network is a clip of $K$ consecutive grayscale video frames, out of which the speaker's face is cropped and scaled to $128 \times 128$ pixels. This results in an input volume of size $128 \times 128 \times K$ scalars, which is then normalized by dividing by maximum pixel intensity and subtracting the mean.

Figure \ref{fig:mouth_vs_fullface} illustrates the importance of allowing the network to learn visual features from the entire face, as opposed to the mouth region only, as widely done in the past. The two lines in the graph represent final network test error as a function of the length $K$ of the clip used as input to the CNN. We tested the values of $K \in \{1,3,5,7,9\}$, while the output $S_i$ always remained the sound features of the center frame. Not surprisingly, the largest gain in performance for both face and mouth regions is when clip length is increased from $K=1$ frame to $K=3$ frames, highlighting the importance of context. The advantage of learning features from the full facial information is also evident, with the best face region error $40\%$ lower than the best mouth region error (both at $K=9$). We hypothesize that this is as result of our CNN using the increased amount of visual information to disambiguate similar mouth movements.

\paragraph*{Sound prediction model}
We use a convolutional neural network (CNN) that takes the aforementioned video clip of size $128 \times 128 \times K$ as input. Our network uses \emph{VGG}-like \cite{vgg} stacks of small $3 \!\times\! 3$ receptive fields in its convolutional layers. The architecture comprises five consecutive $conv3-conv3-maxpool$ blocks consisting of $32-32-64-128-128$ kernels, respectively. These are followed by two fully connected layers with $512$ neurons each. The last layer of our CNN is of size $18$ which corresponds to the size of the sound representation vectors we wish to predict. The network is trained with backpropagation using mean squared error (MSE) loss.

\subsection{Generating a waveform}
\label{ssec:waveform}

Source-filter speech synthesizers such as \cite{klatt1990analysis} use both filter parameters as well as an excitation signal to construct an acoustic signal from LPC features. Predicting excitation parameters is out of the scope of this work, and we therefore use Gaussian white noise as the excitation signal. This produces an \emph{unvoiced} speech signal and results in unnatural sounding speech. Although this method of generating a waveform is relatively simplistic, we found that it worked quite well for speech intelligibility purposes, which is the focus of our work.

\section{Experiments}
\label{experiments}
We applied our speech-reconstruction model to several tasks, and evaluated it with a human listening study.\footnote{Examples of reconstructed speech can be found at \\ \url{http://www.vision.huji.ac.il/vid2speech}}

\paragraph*{Implementation details}
Our network implementation is based on the \emph{Keras} library \cite{chollet2015keras} built on top of \emph{TensorFlow} \cite{tensorflow2015-whitepaper}. Network weights are initialized using the initialization procedure suggested by He \emph{et al.} \cite{He_2015_ICCV}. We use Leaky ReLU \cite{maas2013rectifier} as the non-linear activation function in all layers but the last two, in which we use the hyperbolic tangent (tanh) function. Adam optimizer \cite{kingma2014adam} is used with a learning rate of $0.003$. Dropout \cite{srivastava2014dropout} is used to prevent overfitting, with a rate of $0.25$ after convolutional layers and $0.5$ after fully connected ones. We use mini-batches of $32$ training samples each and stop training when the validation loss stops decreasing (around 80 epochs). Training is done using a single \emph{Nvidia Titan Black} GPU.
We use a cascade-based face detector from \emph{OpenCV} \cite{opencv}, and crop out the mouth region for the comparison in Figure \ref{fig:mouth_vs_fullface} by using a hard-coded mask.
For LPC analysis/resynthesis, as well as excitation generation, we used \emph{pysptk}, a Python wrapper for \emph{Speech Signal Processing Toolkit (SPTK)} \cite{sptk}.

\subsection{GRID corpus}
\label{dataset}

\begin{table}
\centering
\begin{tabular}{@{ }c@{\hspace{3mm}}@{}c@{\hspace{3mm}}@{}c@{\hspace{2mm}}@{}c@{\hspace{2mm}}@{}c@{\hspace{3mm}}@{}c@{ }}
\toprule[1.5pt]
\bf Command&\bf Color&\bf Preposition&\bf Letter&\bf Digit &\bf Adverb	\\
\midrule
bin 	&  blue		& at  	&  A-Z		& 0-9	& again      \\
lay 	&  green 	& by  	&  minus W 	&   	& now      	 \\
place	&  red		& in 	&			& 		& please 	 \\
set		&  white	& with 	& 			& 		& soon		 \\
\bottomrule[1.5pt] \\
\end{tabular}
\caption{GRID sentence grammar.}
\label{tb:grid}
\end{table}

We performed our experiments on the GRID audiovisual sentence corpus \cite{gridcorpus}, a large dataset of audio and video (facial) recordings of $1000$ sentences spoken by $34$ talkers ($18$ male, $16$ female). Each sentence consists of a six word sequence of the form shown in Table \ref{tb:grid}, e.g. ``Place green at H $7$ now".

A total of $51$ different words are contained in the GRID corpus. Videos have a fixed duration of $3$ seconds at a frame rate of $25$ FPS with $720 {\times} 576$ resolution, resulting in sequences comprising $75$ frames. These videos are preprocessed as described in Section \ref{ssec:soundfeatures} before feeding them into the network. The acoustic part of the GRID corpus is used as described in Section \ref{sec:speechrep}.

In order to accurately compare our results with \cite{cornu2015reconstructing}, we performed our experiments on the $1000$ videos of speaker four ($S4$, female) as done there. The training/testing split for each experiment will be described in the following sections.

\subsection{Sound prediction tasks}
\label{ssec:tasks}

\paragraph*{Reconstruction from full dataset}
The first task, proposed by \cite{cornu2015reconstructing}, is designed to examine whether reconstructing audio from visual features can produce intelligible speech. For this task we trained our model on a random $80/20$ train/test split of the $1000$ videos of $S4$ and made sure that all $51$ GRID words were represented in each set. The resulting representation vectors were converted back into waveform using unvoiced excitation, and two different multimedia configurations were constructed: the predicted audio-only and the combination of the original video with reconstructed audio.

\paragraph*{Reconstructing out-of-vocabulary words}
As cited earlier, regression-based models can be used to reconstruct out-of-vocabulary (OOV) words. To test this, we performed the following experiment: The videos in our dataset were sorted according to the digit uttered in each sentence, and our network was trained and tested on five different train/test splits - each with two distinct digits left out of the training set. For example, the network was trained on all sequences with the numbers $1-8$ uttered, and tested only on sequences containing the numbers $9$ and $0$.

\subsection{Evaluating the speech predictions}
\label{ssec:evaluation}

We assessed the intelligibility of the reconstructed speech using a human listening study done using Amazon Mechanical Turk (MTurk). Each job consisted of transcribing one of three types of 3-second clips: audio-only, audio-visual and OOV audio-visual. The listeners were unaware of the differences between the clips. For each clip, they were given the GRID vocabulary and tasked with classifying each reconstructed word into one of its possible options. All together, over $400$ videos containing $38$ distinct sequences were transcribed by $23$ different MTurk workers, which is comparable to the $20$-listener study done by \cite{cornu2015reconstructing}.

\subsection{Results}
\label{ssec:results}

\begin{table}
\centering
\begin{tabular}{lccc}
\toprule[1.5pt]
&  & \multicolumn{2}{c}{\bf Ours} \\
\bf  & \bf \cite{cornu2015reconstructing} & \bf S4 & \bf S2 \\
\midrule
\bf Audio-only 	&  $40.0\%$	&  $\bf 82.6\%$ & - \\
\bf Audio-visual	&  $51.9\%$	&  $\bf 79.9\%$ & $\bf 79\%$  \\
\bottomrule[1.5pt] \\
\end{tabular}
\caption{Our reconstructed speech is significantly more intelligible than the results of \cite{cornu2015reconstructing}. We tested our model on videos from two different speakers in the GRID corpus, $S2$ (male) and $S4$ (female). Randomly guessing a word from each GRID category would result in $19\%$ ``intelligibility".}
\label{tb:lecornu_comparison}
\end{table}

Table \ref{tb:lecornu_comparison} shows the results of our first task, reconstruction from the full dataset, along with a comparison to \cite{cornu2015reconstructing}. Our reconstructed audio is significantly more intelligible than the best results of \cite{cornu2015reconstructing}, as shown by both audio-only and audio-visual tests. The final column shows the result of retraining and testing our model on another speaker from the GRID corpus, speaker two ($S2$, male), whose speech clarity is comparable to $S4$, as reported by \cite{gridcorpus}. We used the same listening test methodology described above, however this time only using combined audio and video.
Examples of original vs. reconstructed LSP coefficients, waveform and spectrogram for this task can be seen in Figure \ref{fig:comparison}.

\begin{table}
\centering
\begin{tabular}{lccc}
\toprule[1.5pt]
\bf  & \bf  OOV & \bf  None out & \bf  Chance					\\
\midrule
\bf Audio-visual 	&  $51.6\%$	&  $93.4\%$  &  $ 10.0\%$      \\
\bottomrule[1.5pt] \\
\end{tabular}
\caption{Out-of-vocabulary (OOV) intelligibility results. We tested this by reconstructing spoken digits which were left out of the training set. Listeners were five times more likely to choose the correct digit than randomly guessing, however only slightly more than half as likely compared to having all digits represented in the training set.}
\label{tb:oov}
\end{table}

Results for the OOV task which appear in Table \ref{tb:oov} were obtained by averaging digit annotation accuracies of the five train/test splits. The fact that human subjects were over five times more likely than chance to choose the correct digit uttered after listening to the reconstructed audio shows that using regression to solve the OOV problem is a promising direction. Moreover, using a larger and more diversified training set vocabulary is likely to significantly increase OOV reconstruction intelligibility. 

\section{Concluding remarks}
\label{sec:conclusion}

This work has proven the feasibility of reconstructing an intelligible audio speech signal from silent videos frames. OOV word reconstruction was also shown to hold promise by modeling automatic speechreading as a regression problem, and using a CNN to automatically learn relevant visual features.

The work described in this paper can serve as a basis for several directions of further research. These include using a less constrained video dataset to show real-world reconstruction viability and generalizing to speaker-independent and multiple speaker reconstruction.


\bibliographystyle{IEEEbib}
\bibliography{vid2speech}

\end{document}